\pdfoutput=1
\documentclass[11pt]{article}

\usepackage[]{emnlp2021}

\usepackage{times}
\usepackage{latexsym}
\usepackage{enumitem}
\usepackage{amsmath}
\usepackage{booktabs}
\usepackage{graphicx}
\usepackage{subcaption}
\usepackage{multirow}
\usepackage{amsfonts}

\usepackage{cancel}

\usepackage[T1]{fontenc}
\usepackage[utf8]{inputenc}

\usepackage{microtype}

\newcommand{\ttoken}{{\scshape TTokens}}
\newcommand{\mtwo}{{\scshape M2-bl}}
\newcommand{\bert}{{\scshape BERT-bl}}
\newcommand{\replay}{{\scshape REPLAY}}
\newcommand{\gem}{{\scshape GEM}}
\newcommand{\replayTT}{{\scshape REPLAY+\ttoken}}
\newcommand{\gemTT}{{\scshape GEM+\ttoken}}
\title{Dynamically Addressing Unseen Rumor via Continual Learning}

\author{
Nayeon Lee \quad Andrea Madotto \quad Yejin Bang \quad Pascale Fung \\
Center for Artificial Intelligence Research (CAiRE) \\
Hong Kong University of Science and Technology \\
\texttt {nyleeaa@connect.ust.hk} \\
}

\begin{document}
\maketitle
\begin{abstract}
Rumors are often associated with newly emerging events, thus, an ability to deal with unseen rumors is crucial for a rumor veracity classification model. Previous works address this issue by improving the model's generalizability, with an assumption that the model will stay unchanged even after the new outbreak of an event.
In this work, we propose an alternative solution to continuously update the model in accordance with the dynamics of rumor domain creations. The biggest technical challenge associated with this new approach is the catastrophic forgetting of previous learnings due to new learnings. 
We adopt continual learning strategies that control the new learnings to avoid catastrophic forgetting and propose an additional strategy that can jointly be used to strengthen the forgetting alleviation.
\end{abstract}

\section{Introduction}
Social media facilitates easy and rapid information sharing that inevitably includes false rumors. A study discovered that false rumors spread farther, faster, and deeper than true rumors~\citep{vosoughi2018spread}. It is, thus, important to devise an automatic system to facilitate the early detection of false rumors before they spread -- rumor veracity classification. Since rumors are often associated with newly emerging breaking news~\citep{zubiaga2018detection}, a crucial technical challenge is to deal with new rumors unseen during the training phase.

\begin{figure}
    \centering
    \includegraphics[width=.8\linewidth]{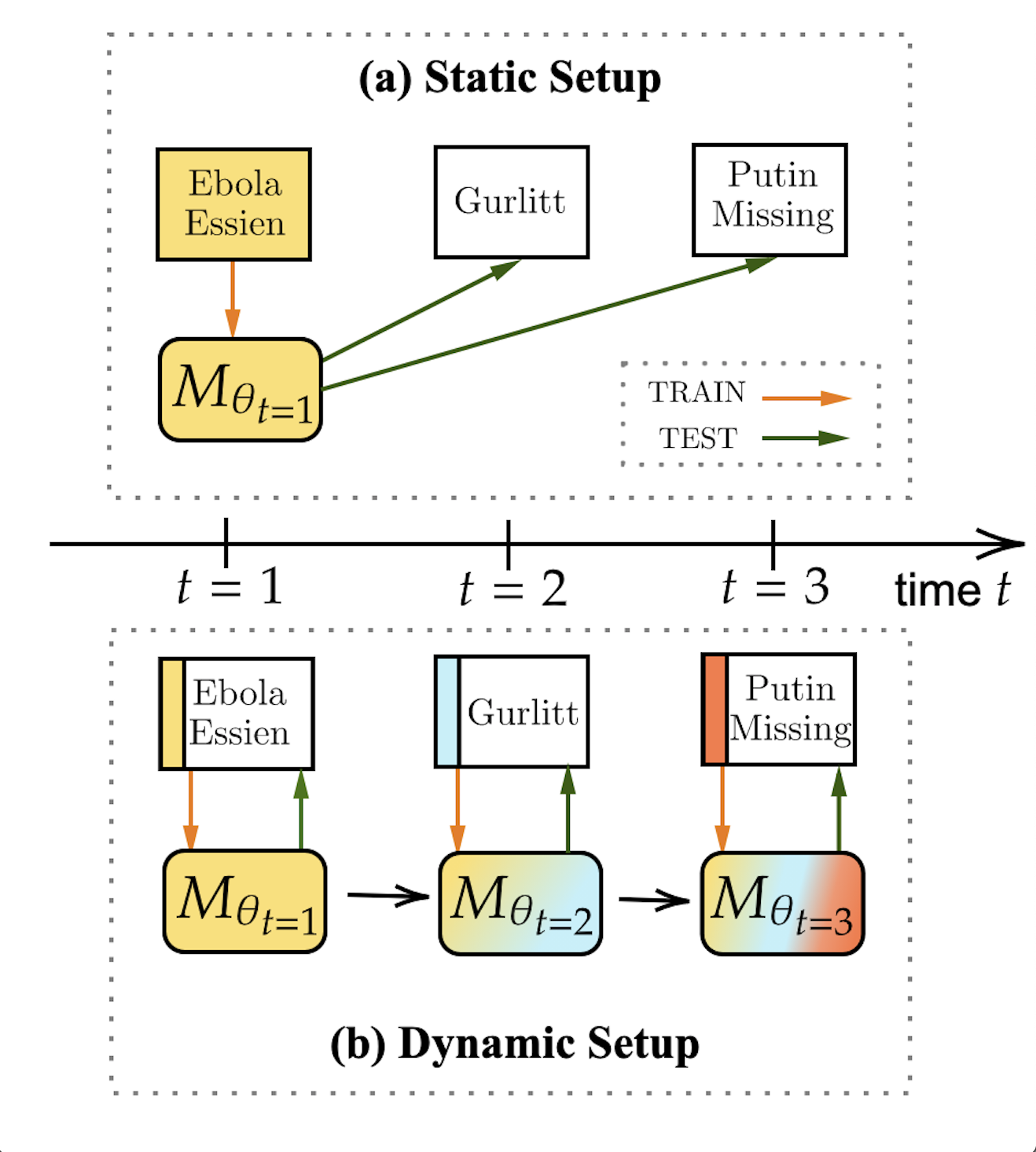}
    \caption{
    % Illustration of rumor veracity classification models learning different domains of rumor through time in (a) static and (b) dynamic setup. 
    Illustration of rumor veracity classification model training procedures in (a) static and (b) dynamic setups. 
    The model in the dynamic setup adapts to unseen domains of rumor through time, whereas the model stays unchanged in the static setup.
    %as expressed with the gradation of colors.
    }
    \label{fig:static_vs_dynamic}
\end{figure}

Previous works~\citep{kochkina2018one,li2019rumor,yu2020coupled,lee2021unifying} have attempted to address this challenge by focusing on the model generalizability in a \textit{static} setup. As illustrated in Figure~\ref{fig:static_vs_dynamic} (a), their objective is to improve the generalizability of the model $M_{\theta_{t=1}}$, trained at $t=1$, to perform well on unseen domains (``Gurlitt'' and ``Putin Missing'') without updating the model.
However, enhancing model generalizability is a hard problem, especially for tasks that always involve the introduction of new topics and vocabulary. Therefore, as an alternative solution, we propose to tackle unseen rumors in a \textit{dynamic} setup by training a classifier that can continuously adapt to newly emerging rumors (Figure~\ref{fig:static_vs_dynamic} (b)). In this way, regardless of the large distributional gap between seen and unseen rumors, we can identify false rumors in a timely manner.

The main challenge of the dynamic setting is the catastrophic forgetting~\cite{mccloskey1989catastrophic} of previously learned domains while new domains are learned.
In this work, to solve the catastrophic forgetting problem, we adopt rehearsal-based continual learning (CL) strategies~\cite{robins1995catastrophic,lopez2017gradient} that use episodic memory from the previously encountered domains to constrain the future learnings, and we additionally propose a simple technique \ttoken~that can jointly be used to further reduce catastrophic forgetting.
Through experiments, we empirically illustrate the benefit of the dynamic setup in comparison to the static setup. To the best of our knowledge, we are the first to address the unseen rumors by continually updating the classifiers with in-domain data. 
% We believe our work suggests a new, promising direction to solve the important challenge associated with fastly evolving rumors. 

% ——————————————— TODO EMNLP submission 때 tackle 하기! ——————————————— 
% Q: how does it can be used in real application? When is this continual learning the most beneficial? (e.g. it can detect timely manner? or it does not need to re train the whole model?) — this question is because of the introduction. intro has good motivation but bit mixed points (timely manner, unseen events)

\section{Methodology}
In In this section, we describe the task definition, the model that serves as a basis of this work, and the training strategies adopted to minimize the catastrophic forgetting and allow effective handling of unseen domains.

\subsection{Task Definition}
\label{section:task_setup_notation}
Rumor veracity classification is the task of identifying whether a given rumor text $X$ is true, false, or unverifiable. More formally, we define a dataset of rumors with their corresponding labels as the set $D=\{(X_i,y_i,\mathrm{Rm})\}_i^N$, where $y\in \{\mathrm{True, False, Unverifiable}\}$, $\mathrm{Rm}$ is the rumor-domain tag and $N$ is the size of the dataset.

The main objective is to train a rumor veracity classification model $M$ that can learn from a stream of rumor-domains through time $t$ without catastrophic forgetting. 
We define the stream of rumor-domains as $\mathcal{S}=\{D_1, \cdots, D_T\}$, where $D_{t}$ represents the dataset of $t$-th rumor domain in the stream and $T$ is the length of the stream. $T$ is also equal to the length of timestamp and the number of rumor domains. At every timestamp, a new rumor domain dataset $D_t$ is used to sequentially train the model $M$, and we denote the model's parameters after training with the rumor domain at time $k$ with $\theta_{t=k}$.

\subsection{Base Model}
Our base model consists of a BERT-base~\cite{devlin2019bert} encoder and a classification head on top. More formally, given the input rumor $X=x_1, \cdots, x_m$, the model computes:
\begin{align}
    H &= \mathrm{BERT}(\mathrm{[CLS]}+X) \label{eq:h} \\ 
   P(y|X) &= \mathrm{Softmax}(W H_{\mathrm{[CLS]}} + b) \label{eq:pxy}
\end{align}
where, $H_{\mathrm{[CLS]}}$ is the embedding of the $\mathrm{[CLS]}$ token and the trainable parameters are $\theta = [W,b]$. 
During training, the encoder layers are frozen and only the classifier parameters $\theta$ are trained using the cross-entropy loss:
\begin{equation}
    L_{\theta_t}(D_t) = - \sum_j^{D_t} \log P(y|X) \label{eq:nll}%(\hat{y_j}) \
\end{equation}

\subsection{Rehearsal-based CL Strategies}
\label{sec:cl1}
Rehearsal-based CL strategies rely on an ``episodic memory'' $\mathcal{M}$ to store previously encountered samples. $\mathcal{M}$ is periodically replayed to avoid catastrophic forgetting and strengthens a connection between past and new knowledge.
% \yeon{is this repetitive to the content in intro? hmm}

\paragraph{REPLAY~\cite{robins1995catastrophic}} 
One simple utilization of the memory $\mathcal{M}$ for CL is to extend the current task data $D_t$, and optimize the models' parameters using $L_{\theta_t}(D_t+\mathcal{M})$.
Basically, it can be viewed as a data-efficient multi-tasking framework that only leverages a small subset of datasets bounded by the size of the memory $\mathcal{M}$.  

\paragraph{Gradient Episodic Memory (GEM)~\cite{lopez2017gradient}}
Another utilization approach is to constrain the gradient updates using current domain samples, so that the loss of the samples in memory $\mathcal{M}$ never increases:
\begin{equation}
    L_{\theta_t}(D_t) \ \text{s.t.} \ L_{\theta_t}(\mathcal{M}) \leq L_{\theta_{t-1}}(\mathcal{M}).
\end{equation}
GEM computes the gradient constraint via a quadratic programming solver that scales with the number of parameters of the model. 

\begin{table}[t]
\centering
\small
\resizebox{\linewidth}{!}{
\begin{tabular}{lcccc}
\toprule
\textbf{Domain} & \textbf{True} & \textbf{False} & \textbf{Unverified} & \textbf{Total} \\ \midrule
Ebola Essien & 0 & 14 & 0 & 14 \\
Gurlitt & 59 & 0 & 2 & 61 \\
Putin missing & 0 & 9 & 117 & 126 \\
Prince Toronto & 0 & 222 & 7 & 229 \\
Germanwings-crash & 94 & 111 & 33 & 238 \\
Ferguson & 10 & 8 & 266 & 284 \\
Charlie Hebdo & 193 & 116 & 149 & 458 \\
Ottawa Shooting & 329 & 72 & 69 & 470 \\
Sydney Siege & 382 & 86 & 54 & 522 \\ \midrule
Total & 1067 & 638 & 697 & 2402 \\ \bottomrule
\end{tabular}
}
\caption{PHEME Dataset Statistics}
\label{table:data_statistics}
\end{table}

\begin{figure*}
\begin{subfigure}{.5\textwidth}
  \centering
      \resizebox{0.8\linewidth}{!}{
        % \begin{tabular}{llcc}
        % \cline{2-4}
        %  & \textit{\textbf{MODEL}} & \textit{\textbf{ACC}} & \textit{\textbf{BWT}} \\ \cline{2-4} 
        %  & \mtwo & 44.3\% & -29.20\% \\
        %  & \bert & 33.77\% & -46.58\% \\ \cline{2-4} 
        % \multirow{4}{*}{\rotatebox{90}{\bert}} & \replay & 70.24\% & -0.62\% \\
        %  & \replayTT & \textbf{75.68\%} & \textbf{4.57\%} \\ \cline{2-4} 
        %  & \gem & 62.01\% & -3.23\% \\
        %  & \gemTT & 68.51\% & 2.86\% \\ \cline{2-4} 
        % \end{tabular}
        \begin{tabular}{lcc}
            \toprule
            MODEL & ACC & BWT \\ \midrule
            \bert & 33.8\% & -46.6\% \\
            \mtwo & 44.3\% & -29.2\% \\
             \midrule
            % +\ttoken & 35.6\% & -40.1\% \\
            \replay & 70.2\% & -0.6\% \\
            \replayTT & \textbf{75.7\%} & \textbf{4.6\%} \\ \midrule
            \gem & 62.0\% & -3.2\% \\ 
            \gemTT & 68.5\% & 2.9\% \\ 
            \bottomrule
        \end{tabular}
      }
  \caption{CL metric at $t=9$. ACC is the average accuracy through time as we do continual learning, and BWT is the measure for catastrophic forgetting; a negative BWT score indicates the existence of catastrophic forgetting, and a positive BWT score indicates the gain in performance for past domain due to future trainings. }
  \label{fig:cl_table}
\end{subfigure}%
\hspace{0.03\textwidth}
\begin{subfigure}{.5\textwidth}
  \centering
  \includegraphics[width=.9\linewidth]{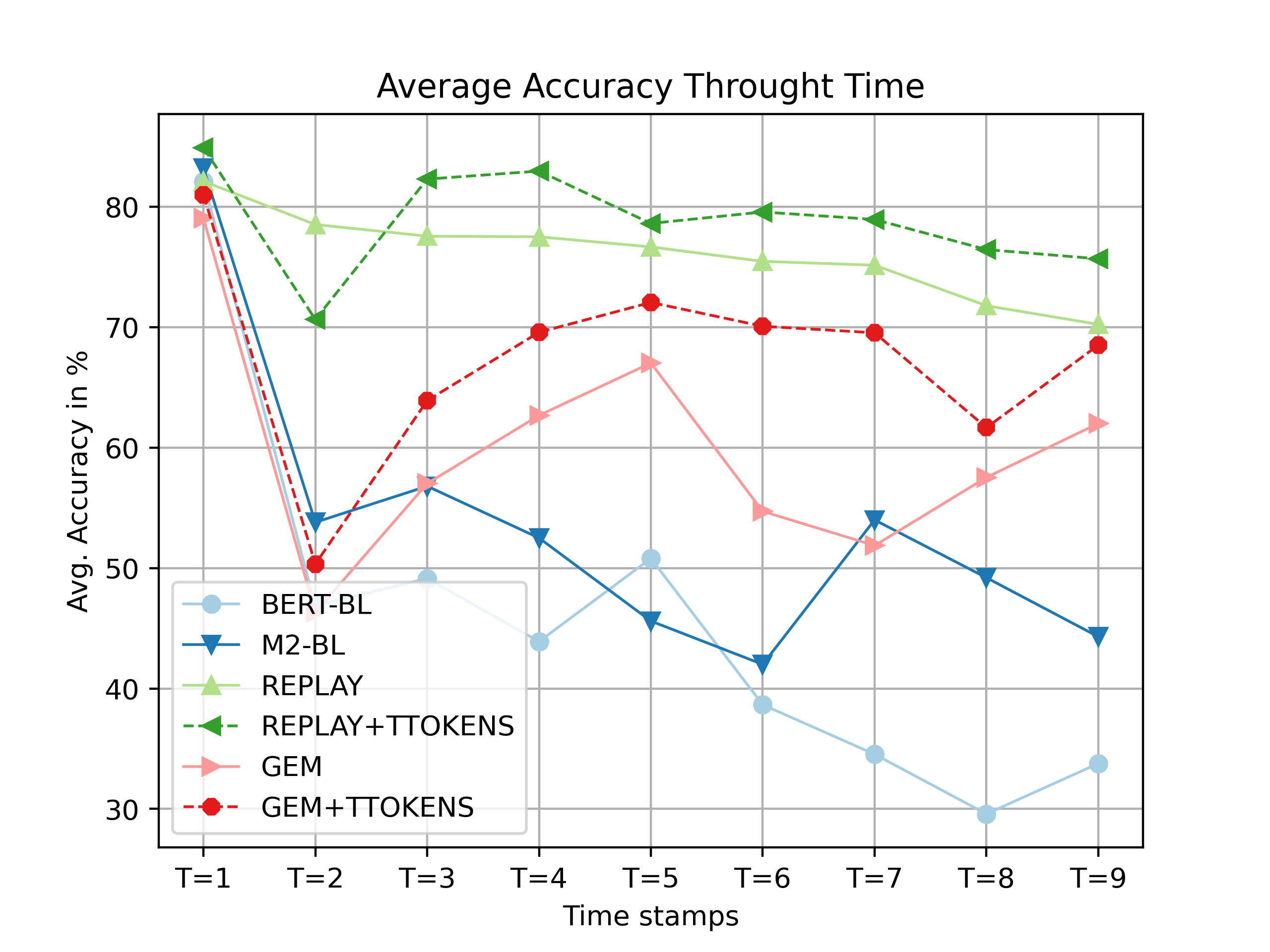}
  \caption{Plot of ACC through time. The ACC scores at $t=9$ are the most important numbers.}
  \label{fig:plot_time}
\end{subfigure}
\caption{Main results. Table a) presents the continual learning performance at $t=9$, and Figure (b) illustrates the plot of ACC through time.}
\label{fig:main_results}
\end{figure*}

\subsection{Task-Specific Tokens (\ttoken)}
\label{sec:cl2}
Various works~\cite{petroni2019language,brown2020language,lee2021towards} illustrated that a context of the input to large pre-trained language models has a huge impact on the outcomes of a model (i.e. representation, generation). In other words, we can leverage this context-dependent characteristic of LM/MLM to intentionally control/distinguish the representation for differing domains.

To apply this strategy, we pre-process the input text $X$ to start with its corresponding rumor-domain tag $\mathrm{Rm}$. 
% Given rumor text sample $X_{i}$ from $D_k$, we concatenate it with rumor specific word/phrase $\mathrm{Rm}_k$. 
Formally, given the $t$-th rumor-domain $\mathrm{Rm}$, Eq.~\ref{eq:h} is replaced with:
\begin{align}
    H &= \mathrm{BERT}(\mathrm{[CLS]}+\mathrm{Rm}+X) %\\ 
    % P(y|X,t) &= \mathrm{Softmax}(W H_{\mathrm{[CLS]}} + b)
\end{align}
% This strategy is applied in the data-processing step thus Eq. \ref{eq:pxy} and \ref{eq:nll} hold. 
This strategy can easily be used together with other CL strategies since it is done in the data-processing step.

% \begin{table*}[t]
% \small
% \centering
% \begin{tabular}{lcccccccccc}
% \toprule
% \multirow{2}{*}{Model} & T=1 & T=2 & T=3 & T=4 & T=5 & T=6 & T=7 & T=8 & \multicolumn{2}{c}{T=9} \\ \cmidrule(lr){2-9}\cmidrule(lr){10-11} 
%  & \multicolumn{8}{c}{ACC} & ACC & BWT \\ \midrule
% \bert & 82.1\% & 46.9\% & 49.1\% & 43.9\% & 50.8\% & 38.7\% & 34.5\% & 29.6\% & 33.8\% & -46.58\% \\ 
% \mtwo & 83.2\%	& 53.8\% & 56.8\% & 52.5\% & 45.6\% & 42.0\% & 54.0\% & 49.2\% & 44.3\% & -29.20\% \\ \midrule
% % +\ttoken & 78.8\% & 46.1\% & 50.9\% & 43.9\% & 31.0\% & 35.1\% & 39.1\% & 31.5\% & 35.6\% & -40.13\% \\ \midrule
% \replay & 82.1\% & \textbf{78.5\%} & 77.6\% & 77.5\% & 76.7\% & 75.5\% & 75.2\% & 71.8\% & 70.2\% & -0.62\% \\
% % \hspace{0.2cm}
% \replayTT & \textbf{84.9\%} & 70.6\% & \textbf{82.3\%} & \textbf{83.0\%} & \textbf{78.6\%} & \textbf{79.6\%} & \textbf{79.0\%} & \textbf{76.5\%} & \textbf{75.7\%} & \textbf{4.57\%} \\ \midrule
% \gem & 79.0\% & 46.3\% & 57.0\% & 62.7\% & 67.0\% & 54.7\% & 51.9\% & 57.5\% & 62.0\% & -3.23\% \\
% \gemTT & 81.0\% & 50.3\% & 63.9\% & 69.6\% & 72.1\% & 70.1\% & 69.5\% & 61.7\% & 68.5\% & 2.86\% \\ \bottomrule
% \end{tabular}
% \caption{ACC through time where number of T is incrementally updated. The ACC scores at T=9 are the most important numbers. \yeon{Update the table - add the M2 result}}
% \label{table:main_cl_acc}
% \end{table*}

\section{Experiments}
\subsection{Dataset and Dynmaic Setup}
We use the rumor veracity classification dataset PHEME, extended and publicly released by \citet{kochkina2018one}. A notable characteristic of PHEME is its categorization by rumor events.
In total, there are nine different events/domains and further details are reported in Table~\ref{table:data_statistics}. 
% Previous works utilize this dataset in a leave-one-event-out cross-validation setting, in which one event is left out for testing while the rest is used for training.

Previous works utilize this dataset in a static setup, where eight domains are combined to be one training set and the remaining one domain is used as a single unseen test domain. 
% To \yeon{support/suit (?)} our dynamic setup, we treat each domain of PHEME to be a separate domain-specific dataset $D_t$, and split them into train/dev/test with the ratio of $0.4/0.1/0.5$.
In this work, our task is carried in a dynamic setup. To incorporate with the dynamic setup, we treat each domain of PHEME to be a separate domain-specific dataset $D_t$, and split them into train/dev/test with the ratio of $0.4/0.1/0.5$.

\subsection{Evaluation Method}
After the model finishes learning about the $k$-th domain, we evaluate its \textit{test} performance on all $T$ domain test sets. The result of this step is a matrix $R\in \mathbb{R}^{T\times T}$, where $R_{i,j}$ is the test classification accuracy of the model on the $j$-th domain 
% $t_j$ 
after observing the last sample from $i$-th domain.
% domain $t_i$. 
Based on this matrix, we compute two CL-specific metrics~\citet{lopez2017gradient}:

\begin{itemize}[noitemsep]
    \item \textbf{Avg. Accuracy (ACC)} is useful for understanding how the performance of the model changes while it is learning new domains. This metric is computed as follows:
    \begin{equation}
        \mathrm{ACC} =\frac{1}{T} \sum_{i=1}^{T} R_{T, i} \label{eq:avgacc}
    \end{equation}
    Note that at the end of the stream, t=9, the Avg. Accuracy (ACC) is exactly the average accuracy of \textit{all} the tasks.
    
    \item \textbf{Backward Transfer (BWT)} is a CL metric used to measure the influence of newly learned tasks to the previously learned ones. This metric is computed as:
    \begin{equation}
        \mathrm{BWT} =\frac{1}{T-1} \sum_{i=1}^{T-1} R_{T, i} - R_{i, i}. \label{eq:bwt}
    \end{equation}
    Notably, negative BWT indicates that the model has catastrophically forgotten the previous tasks.
\end{itemize}

\begin{figure*}
\centering
% \begin{subfigure}{.33\textwidth}
%   \centering
%   \includegraphics[width=.9\linewidth]{trend_over_time.jpeg}
%   \caption{Plot of ACC through time \yeon{TODO: need to update model labels}}
%   \label{fig:temp}
% \end{subfigure}
%
\begin{subfigure}{.45\textwidth}
  \centering
  \includegraphics[width=.9\linewidth]{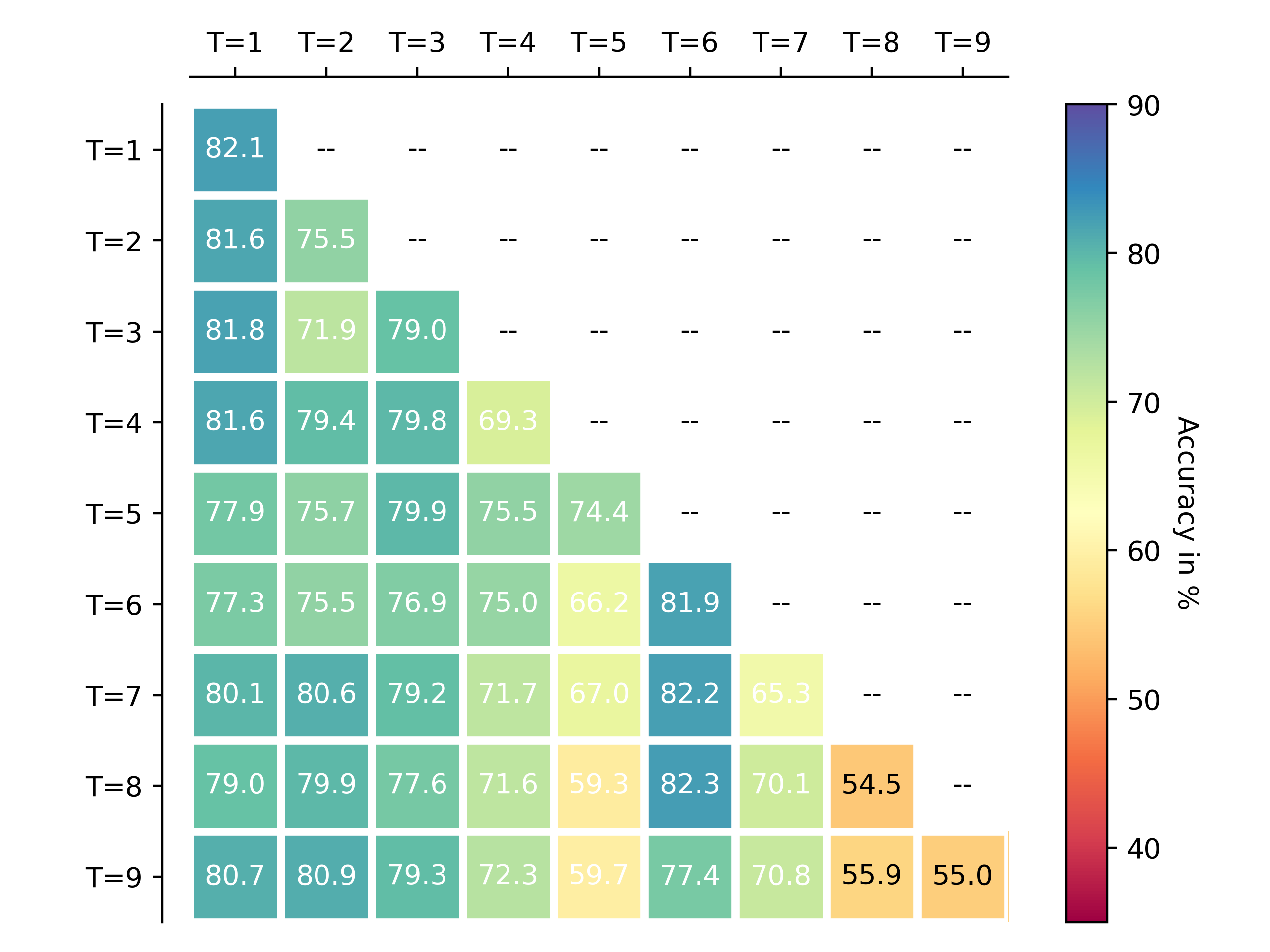}
  \caption{Without \ttoken}
  \label{fig:heatmap_wo_tt}
\end{subfigure}
\begin{subfigure}{.45\textwidth}
  \centering
  \includegraphics[width=.9\linewidth]{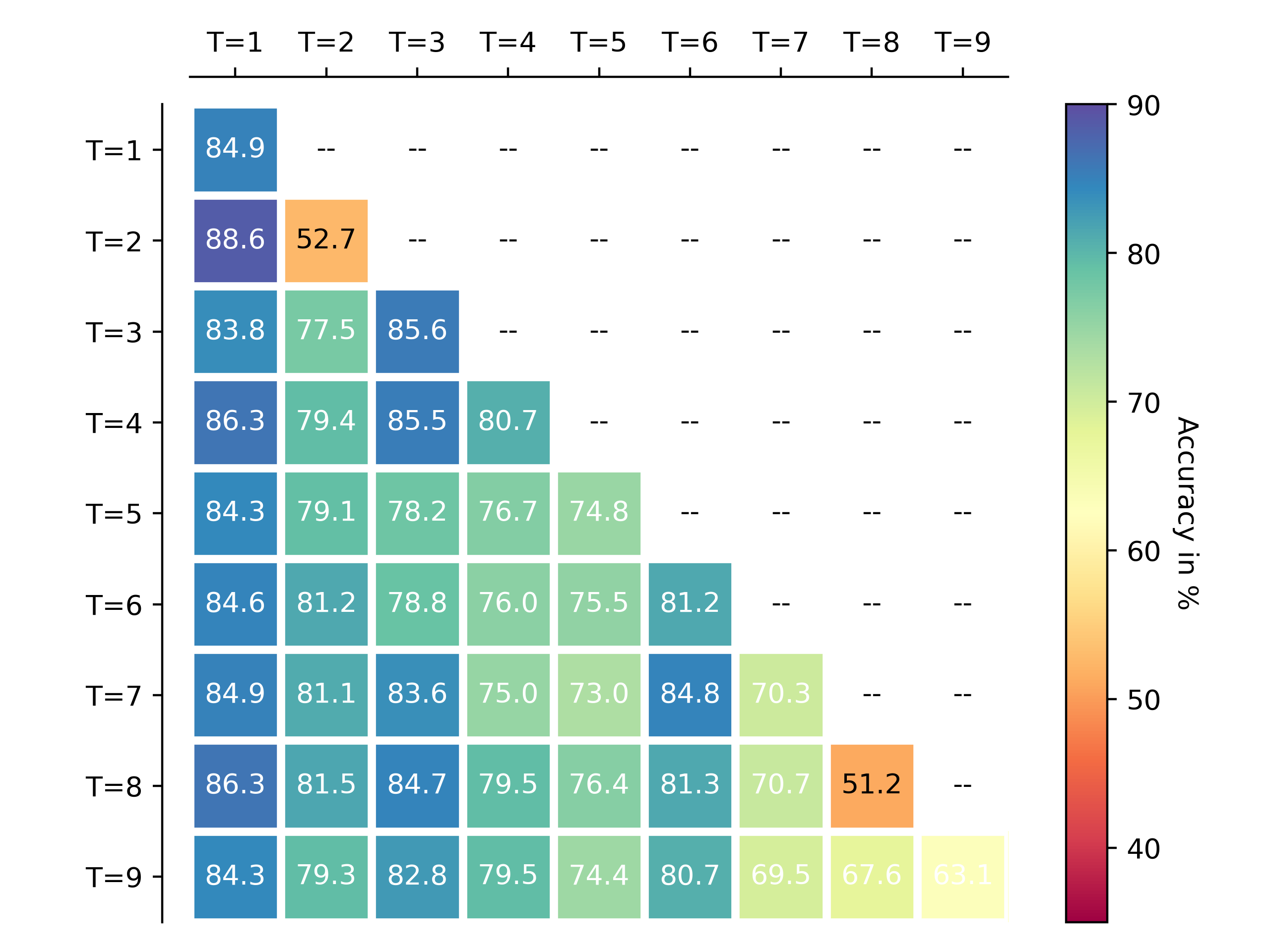}
  \caption{With \ttoken}
  \label{fig:heatmap_w_tt}
\end{subfigure}
\caption{Visualization analysis of model performance through time. Heatmaps of ACC for REPLAY (a) without \ttoken~and (b) with \ttoken. Comparing each column between (a) and (b), columns on (b) with \ttoken~generally show darker shades, which represent better performance.}
\label{fig:visualization}
\end{figure*}

\subsection{Models}
We report two baseline models trained without CL strategy and evaluated in our dynamic setup.
\bert~refers to BERT-base model fine-tuned on PHEME dataset. 
\mtwo~refers to another baseline that fine-tunes the unified misinformation representation~\citep{lee2021unifying} that is shown to be effective in improving the generalizability to unseen domains.

For our proposed models, we have BERT-based classifiers trained with various combinations of CL strategies explained in section \ref{sec:cl1} and \ref{sec:cl2} to evaluate the effectiveness of the adopted strategies on the robustness to unseen domains.

\subsection{Training Details}
We ran all our experiments 3 times with different orders of domains and report the average since there can be a big variance in the overall performance depending on the order of the dataset.
% Each domain dataset is split into train/dev/test using the ratio of $0.45/0.10/0.45$.
For all methods, we used the SGD optimizer~\cite{kiefer1952stochastic} with a mini-batch size of $4$, a learning rate of $0.01$, and the maximum epoch count of $10$. 
For GEM, we used $\lambda=0.5$, which was searched over the $\{0.1,0.3,0.5,0.7,1.0\}$. For REPLAY, we used $M=25$ which is found over $\{25,50,70\}$. These two hyper-parameters were obtained based on the validation-set performance. We ran all experiments with one NVIDIA 2080Ti GPU with 16 GB of memory.

% \begin{figure}[t]
% \centering

% \includegraphics[width=.9\linewidth]{img/REPLAY.png}
% \caption{Heatmap of ACC for  REPLAY without \ttoken}
% \label{fig:heatmap_wo_tt}

% \includegraphics[width=.9\linewidth]{img/REPLAY_TOK.png}
% \caption{Heatmap of ACC for REPLAY with \ttoken}
% \label{fig:heatmap_w_tt}

% \end{figure}

\section{Results and Analysis}
\subsection{Main Results} 
% Our main results are reported in Table~\ref{table:main_cl_acc}.
Our main results are reported in Table~\ref{fig:cl_table} and Figure~\ref{fig:plot_time}.
Firstly, \bert-based classifiers are known to be strong baselines,
% ~\cite{cite,cite}, 
and this is shown to be true from the good performance at $t=1$ ($82\%$ ACC). However, its performance degrades through time, achieving only 33.8\% at the last timestamp ($t=9$). The main cause behind this degradation is, obviously, the effect of catastrophic forgetting, and this is implied from an extremely negative BWT score ($-46.6\%$).
Our second baseline, \mtwo, shows better robustness in comparison to~\bert with $11.5\%$ and $17.4$ gains in the final ACC and BWT scores. However, there still exists a substantial amount of catastrophic forgetting as shown from the halved ACC score and low BWT score ($-29.20\%$) at $t=9$. 

In contrast, all of our dynamic models with CL strategies illustrate better robustness towards catastrophic forgetting. As shown in Figure~\ref{fig:plot_time}, these models start off with a similar performance to the baselines at t=1 but manage to maintain their performance relatively high (i.e., gentler downward slope of the performance).  
Among our models, \replayTT~performs the best with $75.7\%$ ACC in $t=9$, which is almost double of \mtwo. Moreover, it achieves a positive score for the BWT score ($4.57\%$), indicating that it did not only avoid catastrophic forgetting but also managed to transfer additional knowledge from other domains.

\paragraph{Effect of \ttoken} The proposed approach of adding \ttoken~shows a rather impressive effect. Despite its simplicity, we can observe the consistent performance gain from adding it; approximately $5\%$ gains in ACC and BWT for both CL strategies.
Such helpfulness of \ttoken~can also be highlighted from Figure~\ref{fig:plot_time}, where the dotted lines (models \textit{with} \ttoken) are always higher than that the solid lines (models \textit{without} \ttoken). In addition, when we compare the heatmap patterns of the performance between \replay~and \replayTT~(Figure~\ref{fig:heatmap_wo_tt} vs Figure~\ref{fig:heatmap_w_tt}), the colors of \replayTT' heatmap are clearly darker (darker color indicate better performance).

We have two hypotheses for the reasoning behind the success of \ttoken:
1) \ttoken~implicitly served as a signal for the difference in the domain and encouraged the model to learn separate knowledge for each domain. Or,
2) \ttoken~served as a good start ``context'' that helped the LM-based encoder to encode the input to be more separable when necessary -- meaning, input from the same domain to be closer in the vector space than those from the differing domain. 

% \paragraph{Embedding Visualization} \textit{Before} vs \textit{After} adding the \ttoken. \textit{After} visualizations show that the special tokens allow the BERT representations to be more separable between the tasks within the embedding space. As result, reducing the chance of the interference to previously learnt knowledge while optimizing for the new domain.

\section{Related Work}
\paragraph{Rumor Veracity Detection} 
The problem of rumor veracity detection has been actively explored by the NLP community~\citep{kwon2013prominent, ma2017detect, derczynski2017semeval, kochkina2018all, wu-etal-2019-different, li2019rumor, yu2020coupled, yang2020rumor, bang2021model, lee2021unifying}. Some leveraged multi-task learning with stance detection~\citep{kochkina2018all,wu-etal-2019-different} or with other misinformation tasks to enrich the learning~\citep{lee2021unifying}. More recent work utilized advanced architecture such as graph neural network~\citep{yang2020rumor} and hierarchical transformer~\citep{yu2020coupled}. However, these works have only explored rumor veracity detection in a static setup. 

\paragraph{CL in NLP} Continual Learning has been explored for various classification tasks ~\citep{d2019episodic,sprechmann2018memory,wang2020efficient},  generation tasks~\citep{sun2019lamol,hu2020drinking}, sentence encoding~\citep{liu2019continual}, composition language learning~\citep{li2019compositional} and relation learning ~\citep{han2020continual}. Howbeit, to the best of our knowledge, no exploration is done in the rumor verification.
% \paragraph{LAMOL}~\cite{sun2019lamol} generates pseudo-samples of previous tasks for training alongside data for the new task. Here, they use a task-specific token for each task to inform the model to generate pseudo-samples belonging to the specific task. \yeon{our difference - they introduce new token for new task so the vocab size increases everytime. Also, they use it in the context of generation. In our case, we use natural text so our method does not lead to the increase of vocabulary size and the
% embedding weight. Also, we use it for classification task}

\section{Conclusion}
In this work, we proposed the dynamic way of handling unseen rumor domains via continual learning. 
Through experiments, we show that our models with continual learning strategies outperform the strong baselines. Moreover, we highlight the effectiveness of our proposed \ttoken~in further enhancing the overall model performance through time. Aside from \ttoken's effectiveness, its ease of usage makes it more appealing.
We believe our work suggests a new promising direction to solve the important challenge associated with rapidly evolving rumors.

\bibliography{anthology}

\begin{thebibliography}{29}
\expandafter\ifx\csname natexlab\endcsname\relax\def\natexlab#1{#1}\fi

\bibitem[{Bang et~al.(2021)Bang, Ishii, Cahyawijaya, Ji, and
  Fung}]{bang2021model}
Yejin Bang, Etsuko Ishii, Samuel Cahyawijaya, Ziwei Ji, and Pascale Fung. 2021.
\newblock Model generalization on covid-19 fake news detection.
\newblock \emph{arXiv preprint arXiv:2101.03841}.

\bibitem[{Brown et~al.(2020)Brown, Mann, Ryder, Subbiah, Kaplan, Dhariwal,
  Neelakantan, Shyam, Sastry, Askell, Agarwal, Herbert-Voss, Krueger, Henighan,
  Child, Ramesh, Ziegler, Wu, Winter, Hesse, Chen, Sigler, Litwin, Gray, Chess,
  Clark, Berner, McCandlish, Radford, Sutskever, and
  Amodei}]{brown2020language}
Tom Brown, Benjamin Mann, Nick Ryder, Melanie Subbiah, Jared~D Kaplan, Prafulla
  Dhariwal, Arvind Neelakantan, Pranav Shyam, Girish Sastry, Amanda Askell,
  Sandhini Agarwal, Ariel Herbert-Voss, Gretchen Krueger, Tom Henighan, Rewon
  Child, Aditya Ramesh, Daniel Ziegler, Jeffrey Wu, Clemens Winter, Chris
  Hesse, Mark Chen, Eric Sigler, Mateusz Litwin, Scott Gray, Benjamin Chess,
  Jack Clark, Christopher Berner, Sam McCandlish, Alec Radford, Ilya Sutskever,
  and Dario Amodei. 2020.
\newblock \href
  {https://proceedings.neurips.cc/paper/2020/file/1457c0d6bfcb4967418bfb8ac142f64a-Paper.pdf}
  {Language models are few-shot learners}.
\newblock In \emph{Advances in Neural Information Processing Systems},
  volume~33, pages 1877--1901. Curran Associates, Inc.

\bibitem[{d'Autume et~al.(2019)d'Autume, Ruder, Kong, and
  Yogatama}]{d2019episodic}
Cyprien de~Masson d'Autume, Sebastian Ruder, Lingpeng Kong, and Dani Yogatama.
  2019.
\newblock Episodic memory in lifelong language learning.
\newblock \emph{arXiv preprint arXiv:1906.01076}.

\bibitem[{Derczynski et~al.(2017)Derczynski, Bontcheva, Liakata, Procter, Hoi,
  and Zubiaga}]{derczynski2017semeval}
Leon Derczynski, Kalina Bontcheva, Maria Liakata, Rob Procter, Geraldine
  Wong~Sak Hoi, and Arkaitz Zubiaga. 2017.
\newblock Semeval-2017 task 8: Rumoureval: Determining rumour veracity and
  support for rumours.
\newblock In \emph{Proceedings of the 11th International Workshop on Semantic
  Evaluation (SemEval-2017)}, pages 69--76.

\bibitem[{Devlin et~al.(2019)Devlin, Chang, Lee, and
  Toutanova}]{devlin2019bert}
Jacob Devlin, Ming-Wei Chang, Kenton Lee, and Kristina Toutanova. 2019.
\newblock Bert: Pre-training of deep bidirectional transformers for language
  understanding.
\newblock In \emph{Proceedings of the 2019 Conference of the North American
  Chapter of the Association for Computational Linguistics: Human Language
  Technologies, Volume 1 (Long and Short Papers)}, pages 4171--4186.

\bibitem[{Han et~al.(2020)Han, Dai, Gao, Lin, Liu, Li, Sun, and
  Zhou}]{han2020continual}
Xu~Han, Yi~Dai, Tianyu Gao, Yankai Lin, Zhiyuan Liu, Peng Li, Maosong Sun, and
  Jie Zhou. 2020.
\newblock Continual relation learning via episodic memory activation and
  reconsolidation.
\newblock In \emph{Proceedings of the 58th Annual Meeting of the Association
  for Computational Linguistics}, pages 6429--6440.

\bibitem[{Hu et~al.(2020)Hu, Sener, Sha, and Koltun}]{hu2020drinking}
Hexiang Hu, Ozan Sener, Fei Sha, and Vladlen Koltun. 2020.
\newblock Drinking from a firehose: Continual learning with web-scale natural
  language.
\newblock \emph{arXiv preprint arXiv:2007.09335}.

\bibitem[{Kiefer et~al.(1952)Kiefer, Wolfowitz et~al.}]{kiefer1952stochastic}
Jack Kiefer, Jacob Wolfowitz, et~al. 1952.
\newblock Stochastic estimation of the maximum of a regression function.
\newblock \emph{The Annals of Mathematical Statistics}, 23(3):462--466.

\bibitem[{Kochkina et~al.(2018{\natexlab{a}})Kochkina, Liakata, and
  Zubiaga}]{kochkina2018one}
Elena Kochkina, Maria Liakata, and Arkaitz Zubiaga. 2018{\natexlab{a}}.
\newblock \href {https://www.aclweb.org/anthology/C18-1288} {All-in-one:
  Multi-task learning for rumour verification}.
\newblock In \emph{Proceedings of the 27th International Conference on
  Computational Linguistics}, pages 3402--3413, Santa Fe, New Mexico, USA.
  Association for Computational Linguistics.

\bibitem[{Kochkina et~al.(2018{\natexlab{b}})Kochkina, Liakata, and
  Zubiaga}]{kochkina2018all}
Elena Kochkina, Maria Liakata, and Arkaitz Zubiaga. 2018{\natexlab{b}}.
\newblock All-in-one: Multi-task learning for rumour verification.
\newblock In \emph{Proceedings of the 27th International Conference on
  Computational Linguistics}, pages 3402--3413.

\bibitem[{Kwon et~al.(2013)Kwon, Cha, Jung, Chen, and Wang}]{kwon2013prominent}
Sejeong Kwon, Meeyoung Cha, Kyomin Jung, Wei Chen, and Yajun Wang. 2013.
\newblock Prominent features of rumor propagation in online social media.
\newblock In \emph{2013 IEEE 13th International Conference on Data Mining},
  pages 1103--1108. IEEE.

\bibitem[{Lee et~al.(2021{\natexlab{a}})Lee, Bang, Madotto, Khabsa, and
  Fung}]{lee2021towards}
Nayeon Lee, Yejin Bang, Andrea Madotto, Madian Khabsa, and Pascale Fung.
  2021{\natexlab{a}}.
\newblock Towards few-shot fact-checking via perplexity.
\newblock \emph{arXiv preprint arXiv:2103.09535}.

\bibitem[{Lee et~al.(2021{\natexlab{b}})Lee, Li, Wang, Fung, Ma, tau Yih, and
  Khabsa}]{lee2021unifying}
Nayeon Lee, Belinda~Z. Li, Sinong Wang, Pascale Fung, Hao Ma, Wen tau Yih, and
  Madian Khabsa. 2021{\natexlab{b}}.
\newblock \href {http://arxiv.org/abs/2104.05243} {On unifying misinformation
  detection}.

\bibitem[{Li et~al.(2019{\natexlab{a}})Li, Zhang, and Si}]{li2019rumor}
Quanzhi Li, Qiong Zhang, and Luo Si. 2019{\natexlab{a}}.
\newblock Rumor detection by exploiting user credibility information, attention
  and multi-task learning.
\newblock In \emph{Proceedings of the 57th Annual Meeting of the Association
  for Computational Linguistics}, pages 1173--1179.

\bibitem[{Li et~al.(2019{\natexlab{b}})Li, Zhao, Church, and
  Elhoseiny}]{li2019compositional}
Yuanpeng Li, Liang Zhao, Kenneth Church, and Mohamed Elhoseiny.
  2019{\natexlab{b}}.
\newblock Compositional language continual learning.
\newblock In \emph{International Conference on Learning Representations}.

\bibitem[{Liu et~al.(2019)Liu, Ungar, and Sedoc}]{liu2019continual}
Tianlin Liu, Lyle Ungar, and Jo{\~a}o Sedoc. 2019.
\newblock Continual learning for sentence representations using conceptors.
\newblock \emph{arXiv preprint arXiv:1904.09187}.

\bibitem[{Lopez-Paz and Ranzato(2017)}]{lopez2017gradient}
David Lopez-Paz and Marc'Aurelio Ranzato. 2017.
\newblock Gradient episodic memory for continual learning.
\newblock In \emph{Advances in neural information processing systems}, pages
  6467--6476.

\bibitem[{Ma et~al.(2017)Ma, Gao, and Wong}]{ma2017detect}
Jing Ma, Wei Gao, and Kam-Fai Wong. 2017.
\newblock Detect rumors in microblog posts using propagation structure via
  kernel learning.
\newblock In \emph{Proceedings of the 55th Annual Meeting of the Association
  for Computational Linguistics (Volume 1: Long Papers)}, pages 708--717.

\bibitem[{McCloskey and Cohen(1989)}]{mccloskey1989catastrophic}
Michael McCloskey and Neal~J Cohen. 1989.
\newblock Catastrophic interference in connectionist networks: The sequential
  learning problem.
\newblock In \emph{Psychology of learning and motivation}, volume~24, pages
  109--165. Elsevier.

\bibitem[{Petroni et~al.(2019)Petroni, Rockt{\"a}schel, Riedel, Lewis, Bakhtin,
  Wu, and Miller}]{petroni2019language}
Fabio Petroni, Tim Rockt{\"a}schel, Sebastian Riedel, Patrick Lewis, Anton
  Bakhtin, Yuxiang Wu, and Alexander Miller. 2019.
\newblock \href {https://doi.org/10.18653/v1/D19-1250} {Language models as
  knowledge bases?}
\newblock In \emph{Proceedings of the 2019 Conference on Empirical Methods in
  Natural Language Processing and the 9th International Joint Conference on
  Natural Language Processing (EMNLP-IJCNLP)}, pages 2463--2473, Hong Kong,
  China. Association for Computational Linguistics.

\bibitem[{Robins(1995)}]{robins1995catastrophic}
Anthony Robins. 1995.
\newblock Catastrophic forgetting, rehearsal and pseudorehearsal.
\newblock \emph{Connection Science}, 7(2):123--146.

\bibitem[{Sprechmann et~al.(2018)Sprechmann, Jayakumar, Rae, Pritzel, Badia,
  Uria, Vinyals, Hassabis, Pascanu, and Blundell}]{sprechmann2018memory}
Pablo Sprechmann, Siddhant~M Jayakumar, Jack~W Rae, Alexander Pritzel,
  Adria~Puigdomenech Badia, Benigno Uria, Oriol Vinyals, Demis Hassabis, Razvan
  Pascanu, and Charles Blundell. 2018.
\newblock Memory-based parameter adaptation.
\newblock \emph{arXiv preprint arXiv:1802.10542}.

\bibitem[{Sun et~al.(2019)Sun, Ho, and Lee}]{sun2019lamol}
Fan-Keng Sun, Cheng-Hao Ho, and Hung-Yi Lee. 2019.
\newblock Lamol: Language modeling for lifelong language learning.
\newblock In \emph{International Conference on Learning Representations}.

\bibitem[{Vosoughi et~al.(2018)Vosoughi, Roy, and Aral}]{vosoughi2018spread}
Soroush Vosoughi, Deb Roy, and Sinan Aral. 2018.
\newblock The spread of true and false news online.
\newblock \emph{Science}, 359(6380):1146--1151.

\bibitem[{Wang et~al.(2020)Wang, Mehta, P{\'o}czos, and
  Carbonell}]{wang2020efficient}
Zirui Wang, Sanket~Vaibhav Mehta, Barnab{\'a}s P{\'o}czos, and Jaime Carbonell.
  2020.
\newblock Efficient meta lifelong-learning with limited memory.
\newblock \emph{arXiv preprint arXiv:2010.02500}.

\bibitem[{Wu et~al.(2019)Wu, Rao, Jin, Nazir, and Sun}]{wu-etal-2019-different}
Lianwei Wu, Yuan Rao, Haolin Jin, Ambreen Nazir, and Ling Sun. 2019.
\newblock \href {https://doi.org/10.18653/v1/D19-1471} {Different absorption
  from the same sharing: Sifted multi-task learning for fake news detection}.
\newblock In \emph{Proceedings of the 2019 Conference on Empirical Methods in
  Natural Language Processing and the 9th International Joint Conference on
  Natural Language Processing (EMNLP-IJCNLP)}, pages 4644--4653, Hong Kong,
  China. Association for Computational Linguistics.

\bibitem[{Yang et~al.(2020)Yang, Lyu, Tian, Liu, Liu, and
  Zhang}]{yang2020rumor}
Xiaoyu Yang, Yuefei Lyu, Tian Tian, Yifei Liu, Yudong Liu, and Xi~Zhang. 2020.
\newblock Rumor detection on social media with graph structured adversarial
  learning.

\bibitem[{Yu et~al.(2020)Yu, Jiang, Khoo, Chieu, and Xia}]{yu2020coupled}
Jianfei Yu, Jing Jiang, Ling Min~Serena Khoo, Hai~Leong Chieu, and Rui Xia.
  2020.
\newblock \href {https://doi.org/10.18653/v1/2020.emnlp-main.108} {Coupled
  hierarchical transformer for stance-aware rumor verification in social media
  conversations}.
\newblock In \emph{Proceedings of the 2020 Conference on Empirical Methods in
  Natural Language Processing (EMNLP)}, pages 1392--1401, Online. Association
  for Computational Linguistics.

\bibitem[{Zubiaga et~al.(2018)Zubiaga, Aker, Bontcheva, Liakata, and
  Procter}]{zubiaga2018detection}
Arkaitz Zubiaga, Ahmet Aker, Kalina Bontcheva, Maria Liakata, and Rob Procter.
  2018.
\newblock Detection and resolution of rumours in social media: A survey.
\newblock \emph{ACM Computing Surveys (CSUR)}, 51(2):1--36.

\end{thebibliography}
\bibliographystyle{acl_natbib}

\appendix
\newpage
% \section{Example Appendix}
% \label{sec:appendix}

% \begin{table*}[t]
% \small
% \centering
% \begin{tabular}{lcccccccccc}
% \toprule
% \multirow{2}{*}{Model} & t=1 & t=2 & t=3 & t=4 & t=5 & t=6 & t=7 & t=8 & \multicolumn{2}{c}{t=9} \\ \cmidrule(lr){2-9}\cmidrule(lr){10-11} 
%  & \multicolumn{8}{c}{ACC} & ACC & BWT \\ \midrule
% \bert & 82.1\% & 46.9\% & 49.1\% & 43.9\% & 50.8\% & 38.7\% & 34.5\% & 29.6\% & 33.8\% & -46.58\% \\ 
% \mtwo & 83.2\%	& 53.8\% & 56.8\% & 52.5\% & 45.6\% & 42.0\% & 54.0\% & 49.2\% & 44.3\% & -29.20\% \\ \midrule
% % +\ttoken & 78.8\% & 46.1\% & 50.9\% & 43.9\% & 31.0\% & 35.1\% & 39.1\% & 31.5\% & 35.6\% & -40.13\% \\ \midrule
% \replay & 82.1\% & \textbf{78.5\%} & 77.6\% & 77.5\% & 76.7\% & 75.5\% & 75.2\% & 71.8\% & 70.2\% & -0.62\% \\
% % \hspace{0.2cm}
% \replayTT & \textbf{84.9\%} & 70.6\% & \textbf{82.3\%} & \textbf{83.0\%} & \textbf{78.6\%} & \textbf{79.6\%} & \textbf{79.0\%} & \textbf{76.5\%} & \textbf{75.7\%} & \textbf{4.57\%} \\ \midrule
% \gem & 79.0\% & 46.3\% & 57.0\% & 62.7\% & 67.0\% & 54.7\% & 51.9\% & 57.5\% & 62.0\% & -3.23\% \\
% \gemTT & 81.0\% & 50.3\% & 63.9\% & 69.6\% & 72.1\% & 70.1\% & 69.5\% & 61.7\% & 68.5\% & 2.86\% \\ \bottomrule
% \end{tabular}
% \caption{ACC through time. The ACC scores at t=9 are the most important numbers. \yeon{Update the table - add the M2 result}}
% \label{table:main_cl_acc}
% \end{table*}

\end{document}